\pdfoutput=1
\documentclass[11pt]{article}

\usepackage{acl}

\usepackage{times}
\usepackage{latexsym}

\usepackage{amsmath}
\usepackage{multirow}
\usepackage{graphicx}
\usepackage{tablefootnote}
\usepackage{amsfonts}

\usepackage[T1]{fontenc}
\usepackage[utf8]{inputenc}

\usepackage{microtype}

\title{Exploiting Hybrid Semantics of Relation Paths for Multi-hop Question Answering Over Knowledge Graphs}

\author{Zile Qiao$^{1}$, Wei Ye$^{2,}$\footnotemark[2] , Tong Zhang$^{1,3}$, Tong Mo$^{1}$ \\
\bf{Weiping Li$^{1}$, Shikun Zhang$^{2}$}\\
$^1$ School of Software and Microelectronics, Peking University\\
$^2$ National Engineering Research Center for Software Engineering, Peking University \\
$^3$ TencentMT Oteam, China \\
\texttt{\{zileq, wye, zhangtong17, zhangsk\}@pku.edu.cn} \\
\texttt{\{motong, wpli\}@ss.pku.edu.cn}}

\begin{document}

\maketitle
\renewcommand{\thefootnote}{\fnsymbol{footnote}}
\footnotetext[2]{Corresponding authors.}

\begin{abstract}
% Knowledge base question answering (KBQA) remains a great challenge in terms of understanding complex questions via multi-hop reasoning under incomplete KBs. 
Answering natural language questions on knowledge graphs (KGQA) remains a great challenge in terms of understanding complex questions via multi-hop reasoning.
Previous efforts usually exploit large-scale entity-related text corpora or knowledge graph (KG) embeddings as auxiliary information to facilitate answer selection. However, the rich semantics implied in off-the-shelf relation paths between entities is far from well explored. This paper proposes improving multi-hop KGQA by exploiting relation paths' hybrid semantics. Specifically, we integrate explicit textual information and implicit KG structural features of relation paths based on a novel rotate-and-scale entity link prediction framework. Extensive experiments on three existing KGQA datasets demonstrate the superiority of our method, especially in multi-hop scenarios. Further investigation confirms our method's systematical coordination between questions and relation paths to identify answer entities.
\end{abstract}

\section{Introduction}
% Knowledge Base Question Answering (KGQA) is a challenging task to answer natural language questions over a knowledge base (KB)~\cite{Bollacker2008FreebaseAC, Tanon2016FromFT}. 
Answering natural language questions on knowledge graphs (KGQA) is a challenging task~\cite{Bollacker2008FreebaseAC, Tanon2016FromFT}. 
% Early efforts mainly focus on answering a simple question involving a single relational fact~\cite{bordes_large-scale_2015, lan_knowledge_2019}. 
Recent works mainly pay attention to a complex scenario, namely multi-hop KGQA~\cite{sun_open_2018,Hu2018ASF,saxena_improving_2020,Atzeni2021SQALERSQ}, where sophisticated reasoning over multiple edges (or relations) is required to infer the correct answer in the KG~\cite{chen_uhop_2019}.

% \begin{figure}[htbp]
% \centering
% \includegraphics[width=0.482\textwidth]{Intro_case.pdf}
% \caption{Cases that show the relation path between the topic and answer entities. We simplify the name of relative relations in the dataset for easier comprehension.}
% \label{intro-pic}
% \end{figure}

The main challenge of multi-hop KGQA is to understand complicated questions and reason under incomplete KG, usually without supervision signals at the intermediate reasoning steps~\cite{Lan2021KBQASurvey}. One common strategy to alleviate this dilemma is to exploit auxiliary information to enrich knowledge representation. For example, researchers have exploited entity-related textual corpus (e.g., from Wikipedia) as additional nodes in graph-based neural models~\cite{sun_open_2018,sun_pullnet_2019}, or directly encoded them into enhanced entity representations~\cite{Han2020OpenDQ}. A more recent effort, namely EmbedKGQA~\cite{saxena_improving_2020},  leverages implicit yet rich information in KG embeddings to answer complex questions over sparse KG. Unfortunately, the relation paths, which may contain beneficial supplementary information to characterize a candidate target entity for a topic entity in a question, are commonly underutilized. 

% On the contrary, we reveal in this paper that exploiting both explicit textual information and implicit KG embedding features of relation paths could have significantly boosted KBQA performance.

% To the best of our knowledge, there is only one recent work that leverages off-the-shelf relation paths between entities to perform KGQA. To assist the model in answering natural language questions, (reference) uses relation paths as factual evidence from KG. However, we believe that relation paths can be used to provide not only the additional description of entities but also a different view on the corresponding questions. Specifically, we believe that although it is difficult to capture, there is a corresponding relationship between the question and the relation path. These relation paths can also be regarded as a description of the question from another angle which may contain complementary information to questions. 

To the best of our knowledge, 
\citet{Yan2021LargeScaleRL} is the only effort involving exploiting the off-the-shelf relation path information. They use relation paths as simple coarse-grained input features by concatenating their text descriptions. From a more fine-grained and systematic perspective, given a question as a semantic view for the implied relational fact of a <topic entity, target entity> pair, a relation path can serve as another highly-related yet complementary one.

Therefore, we propose coordinating the question view and the relation path view to identify target entities more accurately. To make this idea work, we face two main challenges: 1) how to accurately represent relation paths and 2) how to fuse a relation path representation with the question representation.

% For the first problem, we propose integrating explicit textual semantics and implicit KG embedding features of relation paths. 
For the first problem, we propose exploiting hybrid features of relation paths by integrating both explicit textual semantics and implicit KG embedding features.
Firstly, previous works have shown the merits of introducing entity-related texts~\cite{sun_open_2018,Xiong2019ImprovingQA, sun_pullnet_2019,Han2020OpenDQ}, while we conjecture that relation-related texts (e.g., relation names or descriptions) can potentially offer helpful clues to answer a question. Meanwhile, relation-related texts are naturally available and on a much smaller scale compared with entity-related texts. Therefore, we utilize the explicit text description of a relation path (a relation set) as an extra feature of KGQA models to facilitate target entity selection. 
Secondly, in addition to a text description, a relation also has a KG-based representation (relation embedding) that implicitly contains rich KG structural semantics. Therefore, we introduce RotatE, a KG embedding model that can well support relation composition by entity rotation in the complex vector space~\cite{sun_rotate_2019}. With RotatE, we can synthesize relation path representation by performing simple element-wise multiplication of individual relation embeddings. 
Finally, we characterize beneficial knowledge in candidate relation paths by fusing their structural and textual representations in a question-aware manner, which also facilitates filtering appropriate relation paths semantically consistent with the question among numerous noisy candidates.
    
For the second problem, inspired by~\cite{saxena_improving_2020}, we project the well-designed question-aware mixed representations of relation paths, as well as the question representation, into a rotating entity link prediction framework. However, our pilot experiments showed that the rotating-based link prediction did not yield the robust performance we expected. Further investigation revealed that the modulus of entity embeddings by RotatE mattered. For example, compared with 1-hop answer entities, 2-hop answer entities more significantly differ from their topic entities on modulus. This observation inspires us to match the integrated semantics of questions and relation paths by both entity rotation and entity modulus scaling. After introducing an entity modulus scaling mechanism, we achieve a promising rotate-and-scale prediction framework,  which better coordinate knowledge of questions and relation paths for KGQA.

Extensive experiments on three existing KGQA datasets (WebQuestionSP~\cite{Yih2016TheVO}, ComplexWebQuestions~\cite{Talmor2018TheWA}, and MetaQA~\cite{Zhang2018VariationalRF}) verify the superiority of our method, especially in multi-hop scenarios. Our contributions are as follows:
~\begin{itemize}
\item We propose a KGQA method from a novel perspective of exploiting hybrid features of the off-the-shelf relation paths.

% \item We propose a KGQA method from a novel perspective of exploiting explicit textual and implicit KG structural features of the off-the-shelf relation paths.

\item By systematically fusing explicit \textbf{T}extual information and implicit KG \textbf{E}mbedding features of candidate \textbf{R}elation \textbf{P}aths based on a novel rotate-and-scale KG link prediction framework, our method (\textbf{TERP}) achieves competitive performance on three KGQA datasets, especially in the multi-hop scenario.

\item We reveal that questions and relation paths, as two facets of their corresponding relations between a topic entity and a target entity, are highly-relevant yet complementary information for question answering. 
\end{itemize}

\begin{figure*}[htbp]
\centering
\includegraphics[width=0.96\textwidth]{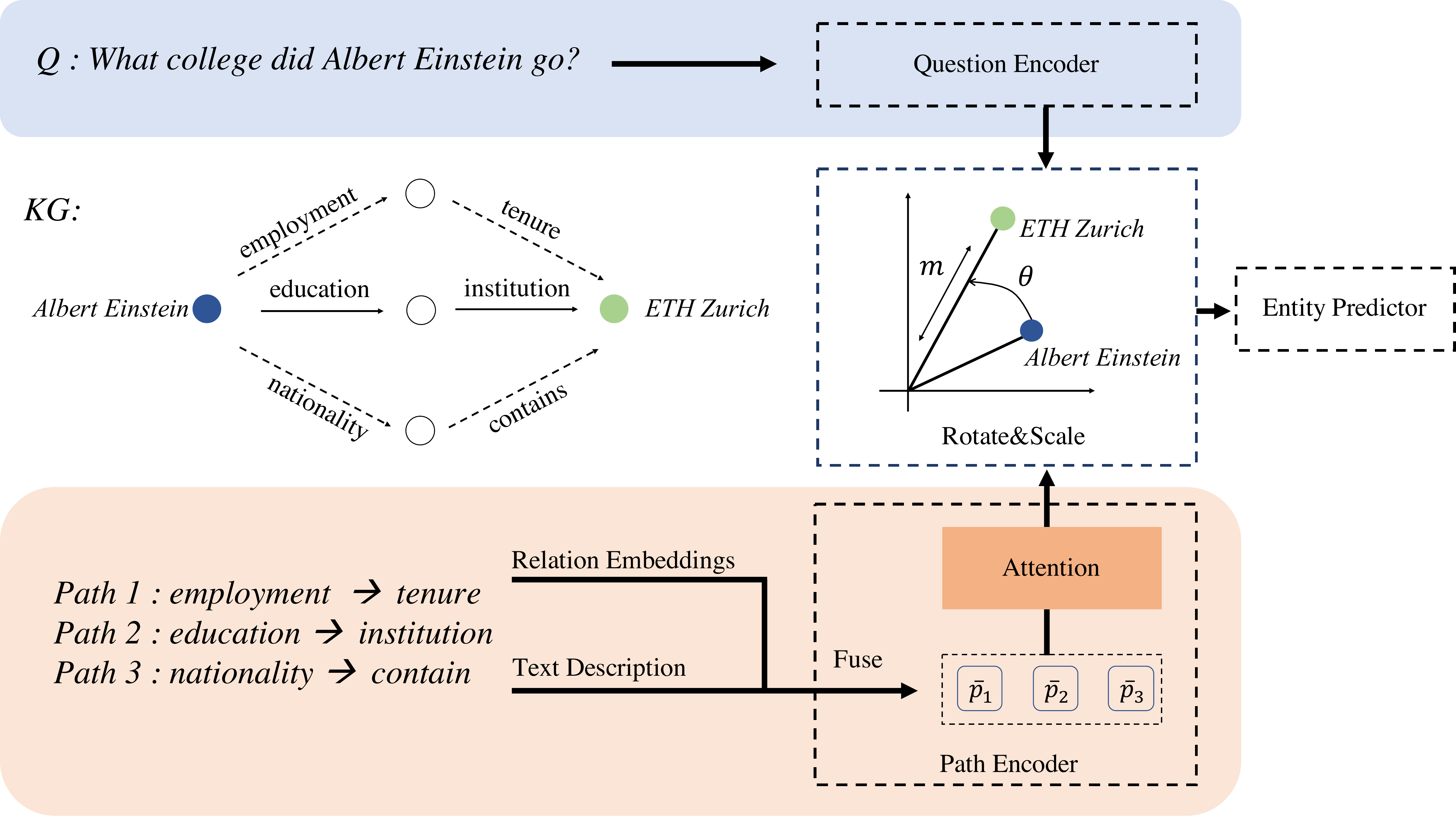}
% \caption{An overview of our TERP model.
% The rotate-and-scale KG link predictor selects answer entities based on entity KG embeddings and question representations. 
% A question encoder is employed to project the textual question $Q$ to a rotate-and-scale transformation $r_q$, matching up with the KG embedding method RotatE. A rotate-and-scale transformation $r$ includes two transformations: a rotation angle $\theta$ and a scaling factor $m$. Additionally, a path encoder is adopted to capture the consistent yet complementary information in implicit relation embeddings (i.e., structural features) and explicit text description (i.e., textual features) of potential relation paths for the question. 

% An attention mechanism from the question to relation paths is used to synthesize essential information among multiple shortest relation paths between a topic entity and an answer entity.
% Finally, a entity predictor scores all candidate entities according to the rotate-and-scale transformations of the question and the relation paths.
% }
\caption{An overview of our TERP model. A question encoder is employed to extract the features from textual question. 
A path encoder is adopted to capture the consistent yet complementary information in implicit relation embeddings (i.e., structural features) and explicit text description (i.e., textual features) of potential relation paths for the question.
Then, the Rotate and Scale mechanism projects the features from question and relation paths to a rotation angle $\theta$ and a scaling factor $m$ , respectively.
Finally, a entity predictor scores all candidate entities.}
\label{main-figure}
\end{figure*}

\section{Problem Statement}
% \subsection{KGQA}

KGQA is the task of factoid question answering over a knowledge graph. A knowledge graph is denoted as $\mathcal{G} \subseteq \mathcal{E} \times \mathcal{R} \times \mathcal{E}$ , where $\mathcal{E}$ is the set of all entities in the KG and $\mathcal{R}$ is the set of all relations. A triple can be formally described as $(h, r, t)$ , where $h,t \in \mathcal{E}$ and $r \in \mathcal{R}$ is the relation between them. Given a natural language question $Q = \left(w_{1}, \ldots, w_{|\mathrm{q}|}\right)$ and a topic entity $h \in \mathcal{E}$, which should be present in the question, the task of KGQA is to extract answer entities $c_a \in \mathcal{E} $ that answer the question $Q$ correctly from $\mathcal{G}$. In practice, we perform entity linking on the question $Q$, producing a set of topic entities $\mathcal{H}$.

\section{Methodology}
% In this section, we present a detailed description of our approach. We start by deﬁning the problem formally and giving an intuitive overview of TERP, 
\subsection{Overview}
In this section, we mathematically present our TERP method in detail.
% Our central intuition behind this work is that the relation path from the topic entity to the answer entity is meaningful for extracting the answer entity. Therefore, we first propose a novel rotate-and-scale KG link prediction framework for KGQA, which leverages RotatE with the capacity to model relation composition. Then, based on the solid KGQA baseline, we explore the explicit textual information and the implicit KG embedding features of the off-the-shelf relation paths between entities, to further capitalize on the relation path information. These hybrid features consisting of text semantics and KG structure semantics are used to synthesize essential knowledge for answer selection based on the attention from the original question to them, hence yielding consistent yet complementary information for the question. Finally, we introduce a two-stage answer selection process to reduce time consumption since there can be too many spurious candidate relation paths in inference.
Our main idea is to coordinate the question view and the relation path view to identify target entities more accurately.
Figure~\ref{main-figure} shows an overview of our approach.
Firstly, we obtain the representations of the entities and relations in KG via a \textbf{KG embedding} module and the representations of questions via a \textbf{Question Encoder}.
% We use a pre-trained RotatE model in our KG embedding module, since it can well support relation composition by entity rotation in the complex vector space~\cite{sun_rotate_2019}. 
Then, we use a \textbf{Path Encoder} to encode the relation path by integrating explicit textual semantics and implicit KG embedding features of relation paths. 
% The representations of question and relation path can be regarded as two view of implied relational fact of a <topic entity, target entity> pair.
An attention mechanism is employed to choose the appropriate relation paths semantically consistent with the question among numerous noisy candidate paths.
A \textbf{Rotate-and-Scale} module projects the representations of the question and the chosen relation paths into the complex space of KG embeddings. 
Finally, an \textbf{Entity Predictor} scores all candidate entities in a link prediction manner.

% As shown in Figure ~\ref{main-figure}, our TERP model includes four main components as follows.

% \paragraph{KG Embedding}
% TERP exploits RotatE KG embeddings, which learn the relation composition by entity rotation in the complex vector space, to facilitate multi-hop KBQA.

% \paragraph{Question Encoder}
% We utilize a RoBERTa-based question encoder to model a textual question as a \emph{rotation} and a \emph{scaling} transformation in a complex space to link the head entity and the tail entity.

% \paragraph{Path Encoder}
% In addition to the question, we further explore the semantic information of the potential relation paths. 
% We first systematically fuse explicit textual information and
% implicit KG embedding features of candidate relation paths. Then we 
% select the appropriate hybrid semantics among multiple relation paths according to the question representation with an attention mechanism.

% % and construct the textual representation of the relation path with a RoBERTa-based path encoder.

% \paragraph{Entity Predictor}
% Finally, an entity predictor scores all candidate entities and delivers the best answer by simultaneously considering the information of the question and relation path.

\subsection{KG Embedding} \label{section:rotate}
We first obtain the representations of entities and relations via a KG Embedding module. To mine the implicit KG structural semantics of relation paths, we adopt RotatE~\cite{sun_rotate_2019} to model the composition of relations.

%  We describe how TERP obtains the representation of entity and relation in this section. 
% To mine the implicit KG structural semantics of relation paths, we need to model the composition of relations. Fortunately, RotatE~\cite{sun_rotate_2019} offers a simple but effective way for that.

RotatE represents entities as complex vectors and relations as rotations in complex vector space. Given $h,t\in \mathcal{E}$ and $r\in \mathcal{R}$, RotatE generates $e_h , e_r, e_t \in \mathbb{C}^{d}$ and deﬁnes a scoring function:
\begin{equation}
    s_{r}(h, t)= \phi(h, r, t) = -\|e_h \circ e_r - e_t\|,
\end{equation}
where $|{e_r}_i|=1$, and $\circ$ denotes the Hadmard (or element-wise) product.
RotatE can model the composition patterns. A relation ${e_r}_{3}=exp({i \theta}_{3})$ is a combination of other two relations ${e_r}_{1}=exp({i \theta}_{1})$ and ${e_r}_{2}=exp({i \theta_{2}})$ if and only if:
\begin{equation}
{e_r}_{3}={e_r}_{1} \circ {e_r}_{2}.
\end{equation}

\subsection{Question Encoder}
% In this section, we introduce the upgraded question encoder matching up with RotatE. 
Following previous work~\cite{saxena_improving_2020, Shi2021TransferNetAE, Atzeni2021SQALERSQ}, the question encoding model aims to embed a natural language question $Q$ to a ﬁxed dimension vector $q \in \mathbb{R}^d$ with a pre-trained language model.
\begin{equation}
    q = \operatorname{Encoder_{avg}}(Q),
\end{equation}
where $\operatorname{avg}$ denotes the average pooling strategy.

\subsection{Path Encoder}
% In this module, we will encode paths. First, we obtain the paths between topic entity and candidate entity through the shortest path algorithm, and directly splice the text information of the relationship to get $\hat{p}$. 
In addition to the question, we further model the explicit and implicit semantics in relation path between a topic entity and a candidate entity with a path encoding module.
Our intuition behind this is that the textual semantics in the relation path, as well as the corresponding composed pre-trained relation embeddings produced by RotatE, are supplementary information to the question.
Considering that the same relation path may have different semantics in different query contexts, we additionally add the question text before the textual description of the relation path.
Formally, given a natural language
question $Q$, a topic entity $h$ and a candidate entity $c$, we can obtain the shortest paths between $h$ and $c$. For a single path ${{r_p}^1},{{r_p}^2},  \cdots{r_p}^k$ , we generate a textual relation path $P_t$, in which every text description of relation in the path is surrounded by special tokens $\mathrm{<r>, </r>}$. We concatenate the textual question $Q$ with the textual relation path $P_t$ and feed them into the encoder to extract the textual features of the explicit relation paths.
% we use special tokens $\mathrm{<r>, </r>}$ surround every text description of relation in the path and splice the question $Q$ in front of them. We name the processed textual path $P_t$.
Meanwhile, we can
obtain the implicit semantics from the embeddings of the relations ${{e_r}_p}^1,{{e_r}_p}^2,  \cdots{{e_r}_p}^k$, which carry structural features learn by RotatE. The hybrid representation $\bar{p}$ of the relation path is produced via fusing both explicit textual representation $p_t$ and implicit KG embedding representation $p_l$ with an $\mathrm{FFN}$:
\begin{equation}
   \begin{split}
    & \bar{p}_t =\operatorname{Encoder_{avg}}(P_t:Q), \\
    & \bar{p}_l = {{e_r}_p}^1\circ {{e_r}_p}^2 \circ \cdots \circ {{e_r}_p}^k, \\
    & \bar{p} = \operatorname{FFN}(\bar{p}_t, \bar{p}_l).
    \end{split} 
\end{equation}

One underlying challenge is that there could be multiple shortest paths between a topic entity and an answer entity. In TERP, we use a scaled dot-product attention mechanism to select the appropriate relation paths that are semantically consistent with the question.
\begin{equation}
    p = \operatorname{Attention}(q \mathbf{W}_1, \mathbf{\bar{p}_t}\mathbf{W}_2, \mathbf{\bar{p}}) ,
\end{equation}
where $\mathbf{\bar{p}_t}$ and $\mathbf{\bar{p}}$ denote two version of representations of all the candidate paths for a given $(h,c)$ pair.
% and the subscript $\mathbf{t}$ denotes textual path.
$\mathbf{W}_1, \mathbf{W}_2$ are learnable matrices. 

% Finally, we handle paths in a similar way to the question. Two independent FFNs are used to model the two transformations. 
% \begin{equation}
%     \begin{split}
%         & \theta_{p}=\operatorname{FFN}(p), \\
%         & m_{p}=\operatorname{FFN}(p), \\
%         & \operatorname{Re}(r_{p}) = m_{p} \circ \cos \left({\theta_{p}}\right), \\
%         & \operatorname{Im}(r_{p}) = m_{p} \circ \sin \left(\theta_{p}\right),
%     \end{split}
%     \label{E2}
% \end{equation}
% where $\theta_{p} \in\mathbb{R}^{d/2}$ and $m_{p} \in\mathbb{R}^{d/2}$ are rotation and scaling in the complex space respectively.

\subsection{Rotate-and-Scale}
However, in our preliminary exploration, directly using RotatE as our KG embedding module do not yield a robust performance.
Further investigation revealed that the underlying reason is that the modulus of entities varies in RotatE. 
If there is an edge between two entities, they will have similar modulus because the modulus of relation representations are fixed to be 1 in RotatE.
However, in a multi-hop KBQA scenario, the multi-hop relation path could amplify the difference between the topic and answer entities, hence it can be challenging to match the answer only by rotation transformation. 
% The representations of question and relation path can be regarded as two view of 

To this end, we propose a rotate-and-scale framework to model the two views of implied relational fact of a <topic entity, target entity> pair as a rotation transformation and a scaling transformation in the complex space.
For a natural language question $Q$, 
% the framework model it as a rotation transformation $\theta_q \in\mathbb{R}^{d}$ and a scaling transformation $m_q \in\mathbb{R}^{d}$ .
two independent feedforward networks ($\operatorname{FFN}$) are used to generate the rotation transformation $\theta_q \in\mathbb{R}^{d}$ and the scaling transformation $m_q \in\mathbb{R}^{d}$:
\begin{equation}
    \begin{split}
        & \theta_{q}=\operatorname{FFN}(q), \\
        & m_{q}=\operatorname{FFN}(q). \\
        % & \operatorname{Re}(r_q) = m_q \circ \cos \left(\theta_{q}\right), \\
        % & \operatorname{Im}(r_q) = m_q \circ \sin \left(\theta_{q}\right).
    \end{split}
    \label{E2}
\end{equation}

Then, we combine the two transformations into the final representation $r_q$ for the question, which contains a real part $\operatorname{Re}$ and an imaginary part $\operatorname{Im}$ in the complex space. 
\begin{equation}
    \begin{split}
        % & \theta_{q}=\operatorname{FFN}(\operatorname{RoBERTa_{avg}}(q)), \\
        % & m_{q}=\operatorname{FFN}(\operatorname{RoBERTa_{avg}}(q)),  \\
        & \operatorname{Re}(r_{q}) = m_{q} \circ \cos \left(\theta_{{q}}\right), \\
        & \operatorname{Im}(r_{q}) = m_{q} \circ \sin \left(\theta_{q}\right).
    \end{split}
    \label{E2}
\end{equation}

In this way, our rotate-and-scale framework can serve as the bridge between 
entity representations of RotatE and representations of textual questions.
We handle relation paths in a same way.
\begin{equation}
    \begin{split}
        & \theta_{p}=\operatorname{FFN}(p), \\
        & m_{p}=\operatorname{FFN}(p), \\
        & \operatorname{Re}(r_{p}) = m_{p} \circ \cos \left({\theta_{p}}\right), \\
        & \operatorname{Im}(r_{p}) = m_{p} \circ \sin \left(\theta_{p}\right),
    \end{split}
    \label{E2}
\end{equation}
where $\theta_{p} \in\mathbb{R}^{d}$ and $m_{p} \in\mathbb{R}^{d}$ are rotation and scaling in the complex space respectively.
As illustrated in Section~\ref{section:Effectiveness}, the rotate-and-scale mechanism improves the performance of KGQA with a large margin.

% We now can select answer entities based on entity embeddings and question representations obtained above, in a way similar to EmbedKGQA. Compared with EmbedKGQA, the unique characteristic of our TERP is that we use RotatE instead of ComplEx and then design the rotate-and-scale mechanism.

\subsection{Entity Predictor}
With the representations produced above, an entity predictor is used to score all candidate entities.
Given a question $Q$, the candidate paths $P$, a topic entity $h \in \mathcal{E}$ and the candidate entity $c \in \mathcal{E}$, the score function is calculated as:
\begin{equation}
   \begin{split}
    & s_{q}(h, c)=-\|e_h \circ r_q - e_c\|, \\
    & s_{p}(h, c)=-\|e_h \circ r_p - e_c\|,
    \end{split} 
\end{equation}
where $s_{q}$ and $s_{p}$ denote the scores from the question view and the relation path view, respectively. The final score is $s = (1-\lambda)s_{q}(h, c) + \lambda s_{p}(h, c)$, where $\lambda$ is a hyper-parameter.
In training, the score $s$ is calculated among $N$ candidate entities sampled from the KG, where $N$ is a hyper-parameter. 

The overall training objective combines the Cross-entropy (\textbf{CE}) loss $\mathcal{L}_{ques}$ and $\mathcal{L}_{path}$ for the $s_{q}$ and $s_{p}$, respectively.
\begin{equation}
    \begin{split}
        \mathcal{L} &= \mathcal{L}_{ques} + \mathcal{L}_{path}  \\
        % &= \operatorname{CE}(s_{p}, targets) \\
        % &+\operatorname{CE}(s_{q}, targets),
        &= \operatorname{CE}(s_{q}, targets)+\operatorname{CE}(s_{p}, targets),
    \end{split}
\end{equation}
where $targets$ denotes the ground truth label. 

\subsection{Inference}
% Following \cite{he_improving_2021, sun_open_2018, sun_pullnet_2019}, we extract a question-specified subgraph with the PPR method for each question and use it as a candidate answer filter during inference. 
% The shortest paths can be directly taken from the subgraph.
To address the challenge that huge numbers of paths may exist, we propose a two-stage inference strategy to reduce the computational cost. At stage 1, given a question $Q$, a topic entity $h$ and all the entities in the question-specified subgraph ${\mathcal{C} \subseteq \mathcal{E}}$, we first compute $s_q(h,c)$ for each $c \in \mathcal{C}$. Then we select top-k candidate entities among them according to $s_q(h,c)$. At stage 2, we compute $s_p(h,c)$ only for the entities recalled in stage 1 and calculate the final score $s$ from them.
For questions with more than one topic entity, we simply average the corresponding $s_q$ and $s_p$ calculated by different topic entities for each candidate $c$.
% remained and compute the final score $s$. 
% Note that the final answers are only selected from the entities remained in stage 2. 

This two-stage answer acquisition strategy can empirically deliver a 15-40$\times$ inference speed-up on different datasets without sacrificing performance.

\begin{table*}[h]
\centering
\begin{tabular}{lcccccc}
\hline
\multicolumn{1}{c}{\multirow{2}{*}{\textbf{Models}}} & \multicolumn{3}{c}{\textbf{MetaQA}}                                                            & \multirow{2}{*}{\textbf{WebQSP}} & \multirow{2}{*}{\textbf{WebQSP-50}} & \multirow{2}{*}{\textbf{CompWebQ}} \\ \cline{2-4}
\multicolumn{1}{c}{}                                 & 1-hop                          & 2-hop                         & 3-hop                         &                                  &                                     &                                    \\ \hline
PullNet*                                             & 97.0                           & 99.9                          & 91.4                          & 68.1                             & 51.9                                & 47.2                               \\
EmbedKGQA                                            & \textbf{97.5} & 98.8                          & 94.8                          & 66.6                             &  54.3$\dagger$                       & 44.7$\dagger$                                  \\
EMQL                                                 & 97.2                           & 98.6                          & 99.1                          & 75.5                             & -                                   & -                                  \\
TransferNet*                                         & 97.5                           & \textbf{100} & \textbf{100} & 71.4                             & -                                   & 48.6                               \\
BERT-KGQA                                           & -                              & -                             & -                             & 71.2                             &  56.7                       & -                                  \\
SQALER*                                               & -                              & 99.9                          & 99.9                          & 76.1                             & 55.2                                & -                                  \\ \hline
TERP(ours)                                           & \textbf{97.5} & 99.4                          & 98.9                          & \textbf{76.8}   & \textbf{57.0}      & \textbf{49.2}     \\ \hline
\end{tabular}
% \caption{Main results on MetaQA, WebQSP, WebQSP-half and CompWebQ. The numbers reported in the table are hits@1.
 
% For a fair comparison, we re-implemented some methods to ensure they use the same WebQSP-50 dataset as ours, corresponding scores are annotated with ``$\dagger$''.}
\caption{Main results on MetaQA, WebQSP, WebQSP-50 and CompWebQ. The numbers reported in the table are hits@1. 
``$\dagger$'' denotes the result of our re-implementation.
Methods that use external corpora are annotated with ``*''.
}
\label{main-results-meta}
\end{table*}

\section{Experimental Settings}
\subsection{Datasets}
We evaluate our model on three widely-used KGQA datasets, MetaQA~\cite{Zhang2018VariationalRF}, WebQuestionsSP~\cite{Yih2016TheVO} datasets, and Complex WebQuestions~\cite{Talmor2018TheWA}. 
% We summarize the statistics of the three datasets in Table~\ref{datasets}.
\paragraph{MetaQA} is a multi-hop KGQA dataset with more than 400k questions, providing a KG with 135k triples, 43k entities and 9 kinds of relations. 
\paragraph{WebQuestionSP(WebQSP)} is a large scale multi-hop KGQA dataset with 4,737 questions. Following \citet{sun_open_2018, sun_pullnet_2019}, we restrict the KG to be a subset of Freebase which contains all facts that are within 2-hops of any entity mentioned in the questions of WebQSP. Then we use the same PPR algorithm as in~\citet{sun_open_2018} to retrieve a subgraph for each question. 
% The overall recall of answers among the subgraphs is 94.0\%.
We further split the testset on WebQSP into 1- and 2-hop sets based on the inferential chain annotation ~\cite{Yih2016TheVO} in the dataset. Note that this split is just for statistics convenience on testsets. During inference, we do not know whether a question is 1-hop or 2-hop, which is different from the MetaQA settings. Following~\citet{sun_open_2018}, we remove half of the triples in the KG to simulate an incomplete KG. We call this setting \textbf{WebQSP-50}. We use the same train/dev/test split as \citet{sun_open_2018}.
\paragraph{ComplexWebQuestions(CompWebQ)} is created by expanding the question entities or adding constraints to the answers in WebQuestionsSP. The questions require up to 4-hops of reasoning on the KG~\cite{he_improving_2021}.We handle CompWebQ in the same way as WebQSP except that we limit each subgraph to a maximum of 2000 entities in CompWebQ. On average, there are 1349 entities in each subgraph and the recall of answers is 78.6\% .

\subsection{Implementation Details}
We use the open source implementation of LibKGE~\cite{libkge} to train the KG embeddings.
Following \citet{saxena_improving_2020}, the pre-trained KG embeddings are frozen for WebQSP and CompWebQ in training, while tuneable for MetaQA.
We use a pre-trained RoBERTa~\cite{Liu2019RoBERTaAR} as the text encoder. 
The size of a mini-batch is set to 10. The learning rate is 3e-5 and we adopt the Adam optimizer with $\beta_2$ = 0.998. The number of candidate entities for WebQSP and CompWebQ is 20000. For MetaQA, we use all entities in KG as candidate entities. Other hyper-parameters are the same as the default RoBERTa-base configuration. The weight $\lambda$ for the entity predictor is 0.6. The number of candidate entities retrieved in stage 1 during inference is empirically set to be 15 for WebQSP and MetaQA, and 30 for CompWebQ.

\subsection{Baselines}
% We compare our TERP with the following baselines: 
\textbf{PullNet}~\cite{sun_pullnet_2019} iteratively retrieves a subgraph from KG to create a question-specific sub-graph and rank the entities by a variant of graph CNN~\cite{Kipf2017SemiSupervisedCW};  \textbf{EmbedKGQA}~\cite{saxena_improving_2020} leverages KG embeddings to perform multi-hop KGQA. It adopts ComplEx~\cite{trouillon_complex_2016} KG embeddings to score the entities;  \textbf{EMQL}~\cite{Sun2020FaithfulEF} leverages query embedding method and uses these embeddings to obtain the answers;  \textbf{TransferNet}~\cite{Shi2021TransferNetAE} leverages free texts retrieved from the textual corpus and pre-defined constrained predicates to perform multi-hop reasoning;  \textbf{BERT-KGQA}~\cite{Yan2021LargeScaleRL} leverages textual information carried by the nodes and edges to perform KGQA. We choose the original version without additional annotated data for a fair comparison;  \textbf{SQALER}~\cite{Atzeni2021SQALERSQ} addresses KGQA by first performing multi-hop reasoning on the KG and then reﬁning the result with logical reasoning.

% \begin{itemize}

%     % \item \textbf{GraftNet}~\cite{sun_open_2018} uses a variant of graph CNN~\cite{Kipf2017SemiSupervisedCW} to perform reasoning over a question-speciﬁc subgraph containing KG facts, entities and sentences from the external text corpora.

%     \item \textbf{PullNet}~\cite{sun_pullnet_2019} learns to retrieve facts and sentences from the external text corpora and KB to create a more relevant question-specific sub-graph. It use the same graph CNN as GraftNet.
    
%     \item \textbf{EmbedKGQA}~\cite{saxena_improving_2020} leverages KG embeddings to perform multi-hop KBQA. It adopts ComplEx~\cite{trouillon_complex_2016} KG embeddings to score the entities.
    
%     % \item \textbf{NSM}~\cite{saxena_improving_2020} uses teacher-student approach for the multi-hop KBQA task. The student network aims to find the correct answer to the query, while the teacher network tries to learn intermediate supervision signals for improving the reasoning capacity of the student network.
    
%     \item \textbf{TransferNet}~\cite{saxena_improving_2020} leverages free texts retrieved from the textual corpus and pre-defined constrained predicates to perform multi-hop reasoning.

% \end{itemize}
\section{Experiment Results}
\subsection{Main Results}

Table~\ref{main-results-meta} shows the performance of the baseline models and our methods on three datasets under different settings. We achieve the best performance on four of six tasks. 
Here we mainly compare our TERP with two lines of works: embedding-based methods (e.g., EmbedKGQA) and path searching methods (e.g., SQALER and TransferNet). 
% We have the following observations.

\textit{Comparison with embedding-based methods.} Except for the similar performance on the MetaQA 1-hop task, TERP significantly outperforms EmbedKGQA on the other tasks. 
% Generally, the superiority of TERP over EmbedKGQA is more evident in more challenging tasks. 
% For example, the performance improvements we achieved on WebQSP and ComWebQ (e.g., +7.9 and +5.7 hits@1, respectively) are more impressive than those on MetaQA tasks (e.g., +0.3 hits@1 for 3-hop). 
The results verify the effectiveness of incorporating relation path information into the link prediction framework.

\textit{Comparison with path searching methods.} Generally, TERP performs better on WebQSP and ComWebQ, while SQALER and TransferNet are more competitive on MetaQA. The possible reason is that the link prediction framework relies on high-quality KG embeddings, consequently being more effective for knowledge graphs of a larger scale. Note that the scale of WebQSP's knowledge graph is far more extensive than that of  MetaQA's (1.8 M v.s. 43 K of entity number, and 6101 v.s. 9 of relation type number). In addition to better trained KG embeddings, the 6101 relation types of WebQSP mean 6101 relation representations, introducing much richer semantics of relations, compared with the 9 relation representations of MetaQA. The difference between these knowledge graphs roughly explains the comparison results. 
Furthermore, the results on more challenging tasks (WebQSP and CompWebQ) verify the effectiveness of integrating explicit textual information and implicit KG structural information in KGQA.

Note that several baselines (EMQL, BERT-KGQA and PullNet) can not accurately fall into the above two categories. Compared with them, TERP also achieves competitive results, e.g., the superior hits@1 scores in 4 of the 6 test sets.

% Compare to the other strong baselines (e.g., EMQL, BERT-KGQA), our TERP shows competitive results, especially on the WebQSP and CompWebQ.
% Furthermore, the unique characteristic of KG embedding-based methods (TERP and EmbedKGQA) is that they relax the requirement on the reachable paths to answer entities, an undesirable constraint imposed by path searching methods. 
% Therefore, it can easily imagine that  NSM and TransferNet can not perform well on incomplete knowledge bases (e.g., WebQSP-half).
% To conclude, by integrating relation path features into KG embedding models, our TERP is more effective for question answering on large-scale knowledge bases, which inevitably have many missing links in practical scenarios. Based on these analysis, we use WebQSP and WebQSP-50 for the following experiments.

\subsection{Effectiveness of Roate-and-Scale Mechanism} \label{section:Effectiveness}

\begin{table*}[ht]
\centering
\begin{tabular}{lcccccc}
\hline
\multicolumn{1}{c}{\multirow{2}{*}{\textbf{Models}}} & \multicolumn{3}{c|}{\textbf{WebQSP}}                                                 & \multicolumn{3}{c}{\textbf{WebQSP-50}}                          \\ \cline{2-7} 
\multicolumn{1}{c}{}                                 & All                 & 1-hop               & \multicolumn{1}{c|}{2-hop}               & All                 & 1-hop               & 2-hop               \\ \hline
\multicolumn{7}{l}{\textbf{w/o path}}                                                                                                                                                                         \\ \hline
w/ ComplEx                                           & 72.1                & 83.6                & \multicolumn{1}{c|}{52.3}                & 53.6                & 63.8                & 36.0                \\
w/ RotatE                                            & 67.5                & 79.7                & \multicolumn{1}{c|}{46.5}                & 49.8                & 60.5                & 31.4                \\
w/ RotatE\&Scale                                     & 74.6                & 84.5                & \multicolumn{1}{c|}{57.5}                & 55.5                & 64.6                & 39.8                \\ \hline
\multicolumn{7}{l}{\textbf{w/ path}}                                                                                                                                                                          \\ \hline
w/ ComplEx                                           & 73.6(+1.5)          & 84.6(+0.9)          & \multicolumn{1}{c|}{54.6(+2.3)}          & 54.6(+1.0)          & 64.2(+0.4)          & 38.0(+2.0)          \\
w/ RotatE                                            & 69.3(+1.8)          & 80.5(+0.8)          & \multicolumn{1}{c|}{50.4(+3.9)}          & 51.0(+0.4)          & 60.6(+0.1)          & 34.2(+2.8)          \\
w/ RotatE\&Scale                                     & \textbf{76.8(+2.2)} & \textbf{84.9(+0.4)} & \multicolumn{1}{c|}{\textbf{62.6(+5.1)}} & \textbf{57.0(+1.5)} & \textbf{65.1(+0.5)} & \textbf{43.1(+3.3)} \\ \hline
\end{tabular}
\caption{Hits@1 on WebQSP datasets in full KG settings (WebQSP) and incomplete KG settings (WebQSP-50). ``All'', ``1-hop'', and ``2-hop'' denote the statistics of 1\&2-, 1-, and 2-hop questions of the same task. 
% Note that the definitions of ``multi-hop'' of WebQSP and MetaQA in terms of evaluation are different. 
\textbf{w/ path} and \textbf{w/o path} denote 
whether the model is equipped with the path encoder. ``w/ ComplEx'' and ``w/ RotatE'' denote the models use ComplEx and RotatE, respectively. ``w/ RotatE\&Scale'' denotes the TERP model with RotatE and the scaling strategy. Numbers in the parentheses denote the hit@1 improvements of \textbf{w/ path} over \textbf{w/o path}.}
\label{main-ablation}
\end{table*}

% One key technical design of our TERP is the rotate-and-scale mechanism to match a potential relation with a question representation. 
To reveal how the rotate-and-scale mechanism helps answer reasoning, we replace it with general ComplEx-based matching and RotatE-based (without scaling) matching, achieving two model variants named \textbf{w/ ComplEx} and \textbf{w/ RotatE} respectively.
% It is noteworthy that \textbf{w/ ComplEx} is generally the same as EmbedKGQA, with its relation matching component removed and then equipped with a PPR filter~\cite{sun_open_2018} used by most previous works.
% This setting can distinguish the real performance effect of the rotate-and-scale mechanism under a unified comparison context.
Table \ref{main-ablation} contains two groups of results corresponding to whether or not hybrid features of relation paths is introduced. The observations on the two groups are generally similar, and here we mainly analyze the results in the first group.

First, \textbf{w/ RotatE} demonstrates a notable performance degradation compared with ~\textbf{w/ ComplEx}, suggesting that simply replacing ComplEx with RotatE in the link prediction-based KGQA framework,  can not satisfy our initial desire to exert relation composition capabilities of RotatE. Second, by incorporating the scaling module into RotatE, \textbf{w/ RotatE\&Scale} surpasses \textbf{w/ ComplEx} with a significant margin. This observation verifies that modulus scaling is necessary to capture relation semantics under the hypothesis of using complex vector rotating to match complex multi-hop questions. Third, the superiority of \textbf{w/ RotatE\&Scale} over \textbf{w/ ComplEx} is more visible on 2-hop questions than that on 1-hop ones, proving that \textbf{w/ RotatE\&Scale} more accurately distinguish relation path semantics.

\subsection{Overall Impacts of Relation Paths' Hybrid Features}

Another characteristic of TERP is using hybrid features of relation paths. In Table~\ref{main-ablation}, the models of the second group are ones with relation path features. By comparing the results of the first group and second group in Table~\ref{main-ablation}, we find 1) incorporating relation path information can consistently improve answering questions of different hops under both complete and incomplete KGs, and 2) the improvements on 2-hop questions surpass that on 1-hop ones by a large margin, verifying the potential of relation path information for multi-hop reasoning.

\subsection{Ablation Study of Relation Paths' Hybrid Features}

We then perform an ablation study on the hybrid features. Table~\ref{ablation2} shows two groups of results corresponding to using only textual representations and only structural representations of relation paths, respectively. ComplEx does not support relation composition, so we only experiment on RotatE and RotatE\&Scale. We have three observations here.

First, both textual and structural features improve model performance, indicating that the two kinds of relation path information benefit answer selection. Second, textual information brings more significant enhancements than structural information. The reasons are two-fold. On the one hand, structural information mainly involves multiplication of relation embedding, which overlaps more with implicit semantics in the link prediction process. On the other hand, textual information provides more complementary knowledge for link prediction, from another modality in a sense. Finally, combining them delivers further improvement, verifying the efficacy of the question-aware fusing process to capture the hybrid semantics.  

\subsection{Collaboration between Questions and Relation Paths}
\begin{table*}[ht]
\centering

\begin{tabular}{lcccccc}
\hline
\multicolumn{1}{c}{\multirow{2}{*}{\textbf{Models}}} & \multicolumn{3}{c|}{\textbf{WebQSP}}                                                      & \multicolumn{3}{c}{\textbf{WebQSP-50}}                                                   \\ \cline{2-7} 
\multicolumn{1}{c}{}                        & \multicolumn{1}{l}{All} & \multicolumn{1}{l}{1-hop} & \multicolumn{1}{l|}{2-hop} & \multicolumn{1}{l}{All} & \multicolumn{1}{l}{1-hop} & \multicolumn{1}{l}{2-hop} \\ \hline
\multicolumn{7}{l}{\textbf{w/ only textual part}}                                                                                                                                                                         \\ \hline
w/ RotatE                                   & 68.4                    & 80.2                      & \multicolumn{1}{c|}{48.2}  & 50.4                    & 60.4                      & 33.2                      \\
w/ RotatE\&Scale                            & 76.1                    & 84.6                      & \multicolumn{1}{c|}{61.6}  & 56.1                    & 64.7                      & 41.4                      \\ \hline
\multicolumn{7}{l}{\textbf{w/ only structural part}}                                                                                                                                                                      \\ \hline
w/ RotatE                                   & 67.8                    & 79.7                      & \multicolumn{1}{c|}{47.3}  & 50.2                    & 60.4                      & 32.7                      \\
w/ RotatE\&Scale                            & 75.3                    & 84.7                      & \multicolumn{1}{c|}{58.9}  & 53.4                    & 61.8                      & 37.9                      \\ \hline
\multicolumn{7}{l}{\textbf{w/ both}}                                                                                                                                                                                      \\ \hline
w/ RotatE                                   & 69.3                    & 80.5                      & \multicolumn{1}{c|}{50.4}  & 51.0                    & 60.6                      & 34.2                      \\
w/ RotatE\&Scale                            & 76.8                    & 84.9                      & \multicolumn{1}{c|}{62.6}  & 57.0                    & 65.1                      & 43.1                      \\ \hline
\end{tabular}
\caption{Ablation results. ``\textbf{w/ only textual part}'' denotes the models that only use the textual features. ``\textbf{w/  only structural part}'' denotes the models that only use the
structural features. ``\textbf{w/  both}'' denotes the models that use both the
structural features and the textual features.}

\label{ablation2}
\end{table*}

\begin{figure}[h]
\centering
\includegraphics[width=0.482\textwidth]{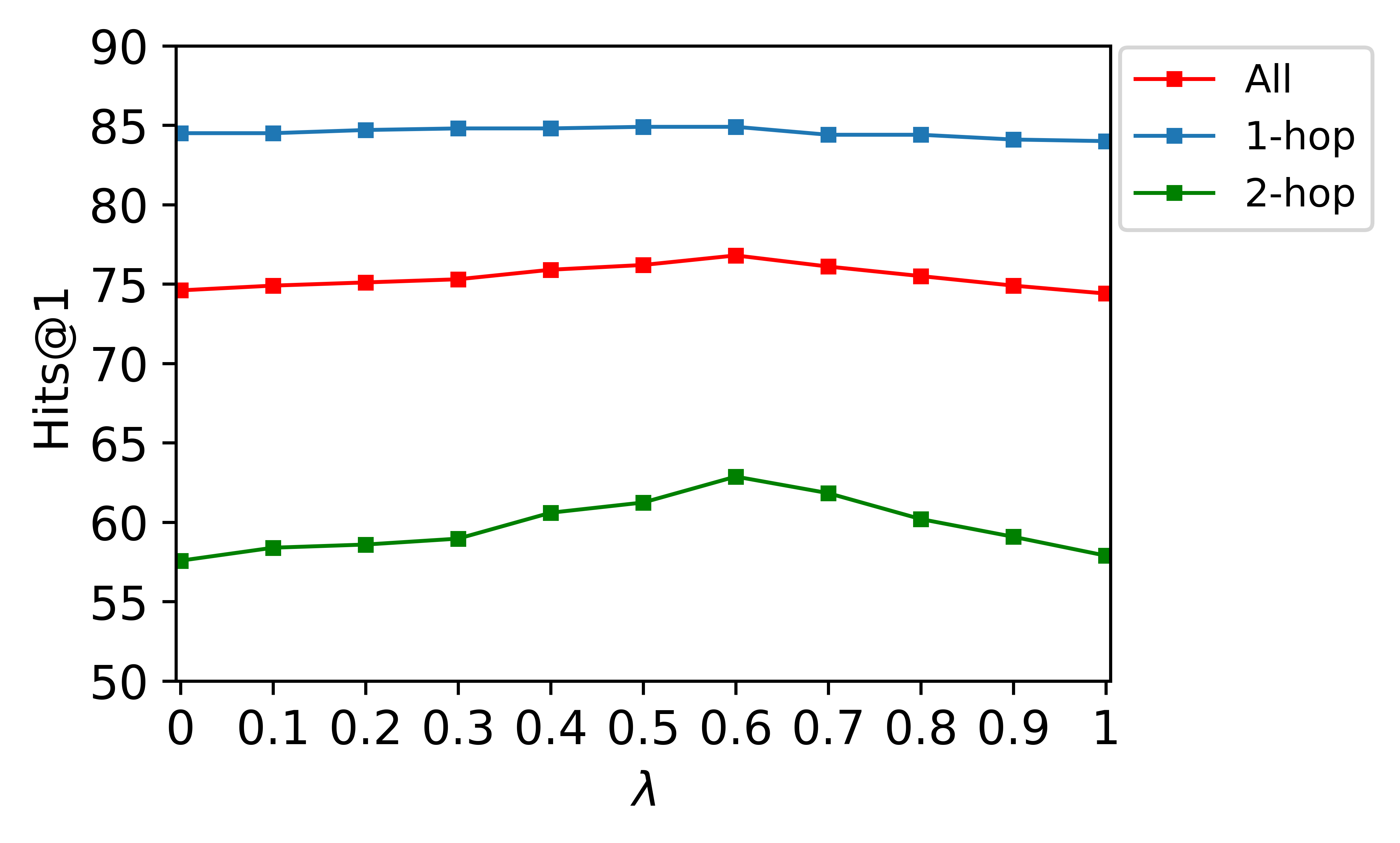}
\caption{ Hits@1 scores for different $\lambda$ on WebQSP. The blue, green, and red lines denotes the testsets with 1-, 2-, and 1\&2-hop questions, respectively.}
\label{weight-loss}
\end{figure}

Considering that exploiting relation paths also introduces many spurious ones, the collaboration of their hybrid features and questions is critical to balance the positive and negative effects. Therefore, we first analyze the impact of the hyper-parameter $\lambda$, which denotes the weighting strategy between predicting scores of questions and relation paths.

In Figure~\ref{weight-loss}, the blue, green, and red polylines show the Hits@1 scores of all 1-hop, 2-hop, and 1\&2-hops questions on WebQSP, respectively. Looking into these three polylines’ trends, we find that our model is best-performed when $\lambda$ is 0.6, indicating the textual information can not either be ignored or overly dependent. In other words, we need to distinguish necessary features under tolerable noises introduced by a set of off-the-shelf relation paths. Another interesting observation is that the upward trend before the peak of the green line (2-hop questions) is more evident than that of the blue line (1-hop questions), though their downward trends after the peak are similar. The reason is that relation path information is more critical for multi-hop reasoning, and our method well characterizes them, hence delivering robust improvements.

\begin{figure}[h]
\centering
\includegraphics[width=0.482\textwidth]{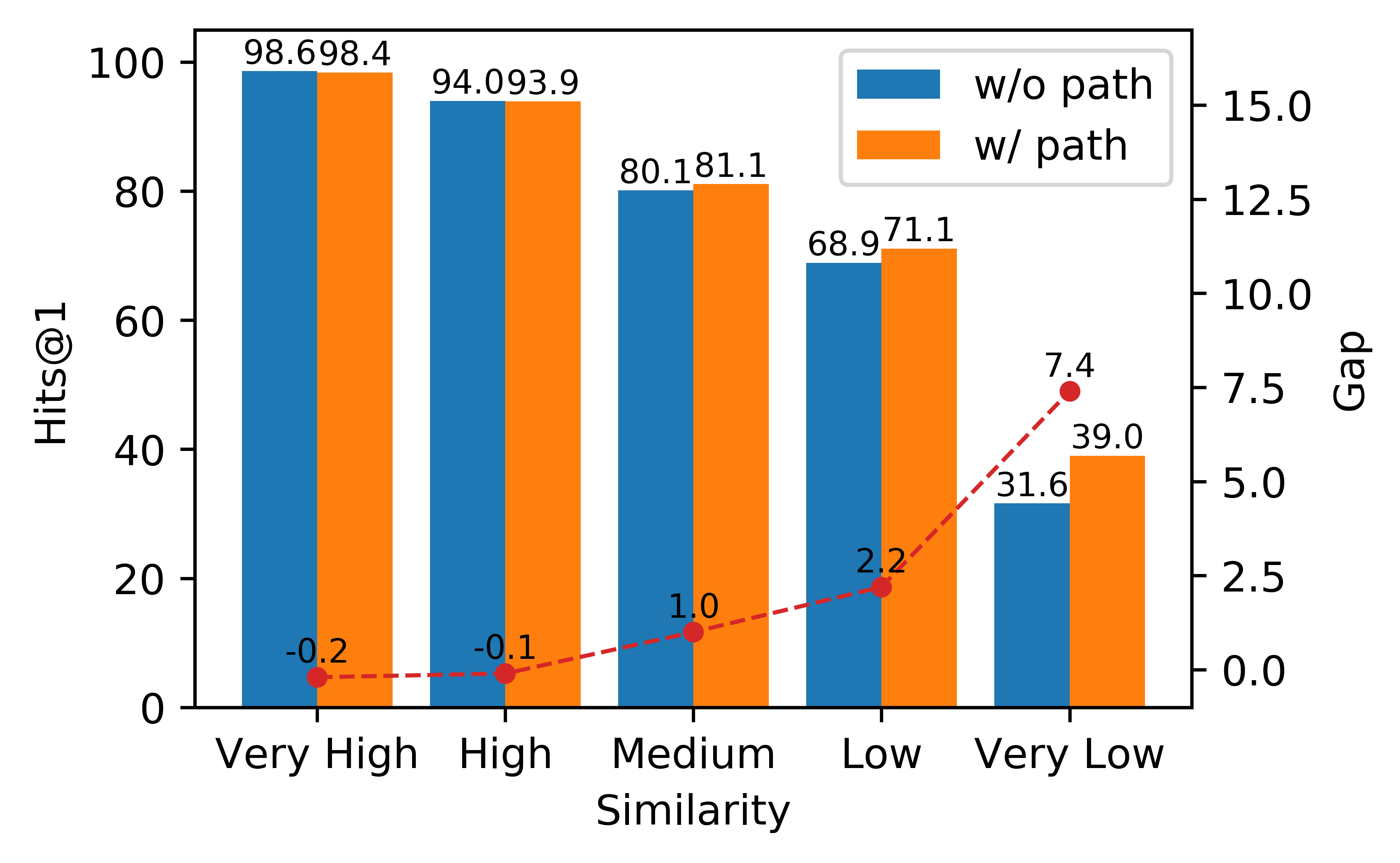}
\caption{Average Hits@1 scores of \textbf{TERP w/o path} and \textbf{TERP w/ path} on sub-testsets of WebQSP with different similarities between question and relation paths. The red line denotes the performance gap between the two compared models.}
\label{similarity}
\end{figure}

To further investigate how relation paths and questions collaborate, we calculate the cosine similarity between relation path text and question text representations for WebQSP.  Since there may be multiple candidate relation paths, the relation path with maximum similarity is selected. We then equally divide data samples in the test set into five groups, based on the cosine similarity scores. The performance of two compared models (\textbf{w/ path} and \textbf{w/o path}) for each group is shown in Figure~\ref{similarity}, from which we observe two interesting trends.

First, model performance degrades as cosine similarity decreases. For example, the hits@1 for ``Very High'' and ``Very Low'' differ enormously (e.g., 98.4 v.s. 39.0 with our full TERP). Intuitively, a question is relatively easy to answer if it is similar to a potential relation path. Otherwise, it is more challenging to find the answer. In other words, the question may not provide enough clues, making question understanding more difficult. 
% That is the reason behind this trend.
Second, the relation path information provides more significant improvement for more difficult questions. Incorporating relation path information may even hinder model performance for the groups of ``Very High'' and ``High'' (e.g., -0.2 and -0.1 hits@1). This is because many relation paths will bring noises but no extra valuable clues. On the contrary, the  hits@1 improvements on ``Medium'', ``Low'', and ``Very Low'' are +1.0, +2.2 and +7.3, respectively. These results clearly demonstrate that relation paths provide complementary information for hard questions, and our method effectively extracts and synthesizes essential features of relation paths. That is where the superiority of our method mainly comes.

% \subsection{Case Study}

% We finally present a case in Table~\ref{case-study} to show how TERP picks the answer entity in real scenarios concretely. Given the question ``What college did Albert Einstein go to,'' which contains the topic entity ``Albert Einstein,'' we present two candidate target entities,``ETH Zurich'' and ``Centerville.''  TERP without the path encoder (\textbf{TERP w/o path}) will wrongly identify ``Centerville'' as the answer. By incorporating the path encoder, however, the full TERP can generate more accurate scores for the two candidate entities with the help of relation paths. The relation path of ``people.person.education $\Rightarrow$En  education.education.institution'' from ``Albert Einstein'' to  ``ETH Zurich'' provides useful clues for finding a ``college,'' and our model correctly identifies this path via attention and hence yields a correct prediction. The case study shows that integrating hybrid features of relation paths with question representations in a link prediction process, not only provides richer semantics to distinguish target entities, but also helps to find the reasoning path to some extent.

\section{Related Work}

% \subsection{KBQA}

There are two categories of KGQA methods commonly known as semantic parsing-based methods and information retrieval-based methods~\cite{Lan2021KBQASurvey}. We mainly focus on the second one. 
% ~\citet{Yang2015KnowledgebasedQA,Bao2016ConstraintBasedQA} proposes extracting a particular sub-graph to answer the question. 
\citet{Miller2016KeyValueMN} proposes to use Memory Networks to learn dense embeddings of the facts present in the KG to perform QA. 
% \citet{Das2017QuestionAO} attempts to use external text to mitigate the KG sparsity problem.
\citet{sun_open_2018,sun_pullnet_2019} create a question-speciﬁc subgraph with entities and sentences from the external text corpora and then use a variant of graph CNN to rank the candidate entities.
Recently, \citet{he_improving_2021} and \citet{Shi2021TransferNetAE} utilize path searching methods to perform KGQA. However, they ignore the information in complete relation path.
\citet{Yan2021LargeScaleRL} leverages relation paths to identify answers, but they only explore the textual form of relation.
In another line of work, \citet{Li2018RepresentationLF} uses TransE~\cite{Bordes2013TranslatingEF} to answer the question, but it cannot work in the scenario of KGQA. EmbedKGQA~\cite{saxena_improving_2020} leverages KG embeddings and projects the question into a link prediction framework.

\section{Conclusion}

We have presented our method for KGQA, which offers a novel perspective of exploiting hybrid features of the off-the-shelf relation paths. We distill essential relation path features by fusing explicit textual information and implicit structural features via a question-aware manner. By projecting a natural language question as well as the obtained hybrid features of candidate relation paths into a novel rotate-and-scale entity link prediction framework, we effectively coordinate question and relation paths to select the answer entity. 
% We also design an efficient two-stage retrieve-and-rank-style candidate selection strategy in inference. 
% Extensive experiments on three KGQA datasets demonstrate the superiority of our method. Furthermore
We reveal that questions and relation paths can be seen as two relevant yet complementary facets of their corresponding relations between a topic entity and a target entity.
% We wish this finding could inspire more future work on exploring relation path information for KGQA.

\section*{Acknowledgements}
We thank anonymous reviewers for their valuable comments. This research was supported by the National Key Research and Development Program of China (No. 2018YFB1403002).

\bibliography{anthology,custom}

\begin{thebibliography}{26}
\expandafter\ifx\csname natexlab\endcsname\relax\def\natexlab#1{#1}\fi

\bibitem[{Atzeni et~al.(2021)Atzeni, Bogojeska, and
  Loukas}]{Atzeni2021SQALERSQ}
Mattia Atzeni, Jasmina Bogojeska, and Andreas Loukas. 2021.
\newblock Sqaler: Scaling question answering by decoupling multi-hop and
  logical reasoning.
\newblock In \emph{NeurIPS}.

\bibitem[{Bollacker et~al.(2008)Bollacker, Evans, Paritosh, Sturge, and
  Taylor}]{Bollacker2008FreebaseAC}
Kurt~D. Bollacker, Colin Evans, Praveen~K. Paritosh, Tim Sturge, and Jamie
  Taylor. 2008.
\newblock Freebase: a collaboratively created graph database for structuring
  human knowledge.
\newblock In \emph{SIGMOD Conference}.

\bibitem[{Bordes et~al.(2013)Bordes, Usunier, Garc{\'i}a-Dur{\'a}n, Weston, and
  Yakhnenko}]{Bordes2013TranslatingEF}
Antoine Bordes, Nicolas Usunier, Alberto Garc{\'i}a-Dur{\'a}n, Jason Weston,
  and Oksana Yakhnenko. 2013.
\newblock Translating embeddings for modeling multi-relational data.
\newblock In \emph{NIPS}.

\bibitem[{Broscheit et~al.(2020)Broscheit, Ruffinelli, Kochsiek, Betz, and
  Gemulla}]{libkge}
Samuel Broscheit, Daniel Ruffinelli, Adrian Kochsiek, Patrick Betz, and Rainer
  Gemulla. 2020.
\newblock \href {https://www.aclweb.org/anthology/2020.emnlp-demos.22}
  {{L}ib{KGE} - {A} knowledge graph embedding library for reproducible
  research}.
\newblock In \emph{Proceedings of the 2020 Conference on Empirical Methods in
  Natural Language Processing: System Demonstrations}, pages 165--174.

\bibitem[{Chen et~al.(2019)Chen, Chang, Chen, Nayak, and Ku}]{chen_uhop_2019}
Zi-Yuan Chen, Chih-Hung Chang, Yi-Pei Chen, Jijnasa Nayak, and Lun-Wei Ku.
  2019.
\newblock \href {http://arxiv.org/abs/1904.01246} {{UHop}: {An}
  {Unrestricted}-{Hop} {Relation} {Extraction} {Framework} for
  {Knowledge}-{Based} {Question} {Answering}}.
\newblock \emph{arXiv:1904.01246 [cs]}.
\newblock ArXiv: 1904.01246.

\bibitem[{Han et~al.(2020)Han, Cheng, and Wang}]{Han2020OpenDQ}
Jiale Han, Bo~Cheng, and Xu~Wang. 2020.
\newblock Open domain question answering based on text enhanced knowledge graph
  with hyperedge infusion.
\newblock In \emph{FINDINGS}.

\bibitem[{He et~al.(2021)He, Lan, Jiang, Zhao, and Wen}]{he_improving_2021}
Gaole He, Yunshi Lan, Jing Jiang, Wayne~Xin Zhao, and Ji-Rong Wen. 2021.
\newblock \href {https://doi.org/10.1145/3437963.3441753} {Improving
  {Multi}-hop {Knowledge} {Base} {Question} {Answering} by {Learning}
  {Intermediate} {Supervision} {Signals}}.
\newblock \emph{Proceedings of the 14th ACM International Conference on Web
  Search and Data Mining}, pages 553--561.
\newblock ArXiv: 2101.03737.

\bibitem[{Hu et~al.(2018)Hu, Zou, and Zhang}]{Hu2018ASF}
Sen Hu, Lei Zou, and Xinbo Zhang. 2018.
\newblock A state-transition framework to answer complex questions over
  knowledge base.
\newblock In \emph{EMNLP}.

\bibitem[{Kipf and Welling(2017)}]{Kipf2017SemiSupervisedCW}
Thomas Kipf and Max Welling. 2017.
\newblock Semi-supervised classification with graph convolutional networks.
\newblock \emph{ArXiv}, abs/1609.02907.

\bibitem[{Lan et~al.(2021)Lan, He, Jiang, Jiang, Zhao, and
  Wen}]{Lan2021KBQASurvey}
Yunshi Lan, Gaole He, Jinhao Jiang, Jing Jiang, Wayne~Xin Zhao, and Ji{-}Rong
  Wen. 2021.
\newblock \href {https://doi.org/10.24963/ijcai.2021/611} {A survey on complex
  knowledge base question answering: Methods, challenges and solutions}.
\newblock In \emph{Proceedings of the Thirtieth International Joint Conference
  on Artificial Intelligence, {IJCAI} 2021, Virtual Event / Montreal, Canada,
  19-27 August 2021}, pages 4483--4491. ijcai.org.

\bibitem[{Li et~al.(2018)Li, Zhang, and Li}]{Li2018RepresentationLF}
Dingcheng Li, Jingyuan Zhang, and Ping Li. 2018.
\newblock Representation learning for question classification via topic sparse
  autoencoder and entity embedding.
\newblock \emph{2018 IEEE International Conference on Big Data (Big Data)},
  pages 126--133.

\bibitem[{Liu et~al.(2019)Liu, Ott, Goyal, Du, Joshi, Chen, Levy, Lewis,
  Zettlemoyer, and Stoyanov}]{Liu2019RoBERTaAR}
Yinhan Liu, Myle Ott, Naman Goyal, Jingfei Du, Mandar Joshi, Danqi Chen, Omer
  Levy, Mike Lewis, Luke Zettlemoyer, and Veselin Stoyanov. 2019.
\newblock Roberta: A robustly optimized bert pretraining approach.
\newblock \emph{ArXiv}, abs/1907.11692.

\bibitem[{Miller et~al.(2016)Miller, Fisch, Dodge, Karimi, Bordes, and
  Weston}]{Miller2016KeyValueMN}
Alexander~H. Miller, Adam Fisch, Jesse Dodge, Amir-Hossein Karimi, Antoine
  Bordes, and Jason Weston. 2016.
\newblock Key-value memory networks for directly reading documents.
\newblock In \emph{EMNLP}.

\bibitem[{Saxena et~al.(2020)Saxena, Tripathi, and
  Talukdar}]{saxena_improving_2020}
Apoorv Saxena, Aditay Tripathi, and Partha Talukdar. 2020.
\newblock \href {https://doi.org/10.18653/v1/2020.acl-main.412} {Improving
  {Multi}-hop {Question} {Answering} over {Knowledge} {Graphs} using
  {Knowledge} {Base} {Embeddings}}.
\newblock In \emph{Proceedings of the 58th {Annual} {Meeting} of the
  {Association} for {Computational} {Linguistics}}, pages 4498--4507, Online.
  Association for Computational Linguistics.

\bibitem[{Shi et~al.(2021)Shi, Cao, Hou, Li, and Zhang}]{Shi2021TransferNetAE}
Jiaxin Shi, Shulin Cao, Lei Hou, Juan-Zi Li, and Hanwang Zhang. 2021.
\newblock Transfernet: An effective and transparent framework for multi-hop
  question answering over relation graph.
\newblock In \emph{EMNLP}.

\bibitem[{Sun et~al.(2020)Sun, Arnold, Bedrax-Weiss, Pereira, and
  Cohen}]{Sun2020FaithfulEF}
Haitian Sun, Andrew~O. Arnold, Tania Bedrax-Weiss, Fernando Pereira, and
  William~W. Cohen. 2020.
\newblock Faithful embeddings for knowledge base queries.
\newblock \emph{arXiv: Learning}.

\bibitem[{Sun et~al.(2019{\natexlab{a}})Sun, Bedrax-Weiss, and
  Cohen}]{sun_pullnet_2019}
Haitian Sun, Tania Bedrax-Weiss, and William~W. Cohen. 2019{\natexlab{a}}.
\newblock \href {http://arxiv.org/abs/1904.09537} {{PullNet}: {Open} {Domain}
  {Question} {Answering} with {Iterative} {Retrieval} on {Knowledge} {Bases}
  and {Text}}.
\newblock \emph{arXiv:1904.09537 [cs]}.
\newblock ArXiv: 1904.09537.

\bibitem[{Sun et~al.(2018)Sun, Dhingra, Zaheer, Mazaitis, Salakhutdinov, and
  Cohen}]{sun_open_2018}
Haitian Sun, Bhuwan Dhingra, Manzil Zaheer, Kathryn Mazaitis, Ruslan
  Salakhutdinov, and William~W. Cohen. 2018.
\newblock \href {http://arxiv.org/abs/1809.00782} {Open {Domain} {Question}
  {Answering} {Using} {Early} {Fusion} of {Knowledge} {Bases} and {Text}}.
\newblock \emph{arXiv:1809.00782 [cs]}.
\newblock ArXiv: 1809.00782.

\bibitem[{Sun et~al.(2019{\natexlab{b}})Sun, Deng, Nie, and
  Tang}]{sun_rotate_2019}
Zhiqing Sun, Zhi-Hong Deng, Jian-Yun Nie, and Jian Tang. 2019{\natexlab{b}}.
\newblock \href {http://arxiv.org/abs/1902.10197} {{RotatE}: {Knowledge}
  {Graph} {Embedding} by {Relational} {Rotation} in {Complex} {Space}}.
\newblock \emph{arXiv:1902.10197 [cs, stat]}.
\newblock ArXiv: 1902.10197.

\bibitem[{Talmor and Berant(2018)}]{Talmor2018TheWA}
Alon Talmor and Jonathan Berant. 2018.
\newblock The web as a knowledge-base for answering complex questions.
\newblock In \emph{NAACL}.

\bibitem[{Tanon et~al.(2016)Tanon, Vrande, Schaffert, Steiner, and
  Pintscher}]{Tanon2016FromFT}
Thomas~Pellissier Tanon, Denny Vrande, Sebastian Schaffert, Thomas Steiner, and
  Lydia Pintscher. 2016.
\newblock From freebase to wikidata: The great migration.
\newblock \emph{Proceedings of the 25th International Conference on World Wide
  Web}.

\bibitem[{tau Yih et~al.(2016)tau Yih, Richardson, Meek, Chang, and
  Suh}]{Yih2016TheVO}
Wen tau Yih, Matthew Richardson, Christopher Meek, Ming-Wei Chang, and Jina
  Suh. 2016.
\newblock The value of semantic parse labeling for knowledge base question
  answering.
\newblock In \emph{ACL}.

\bibitem[{Trouillon et~al.(2016)Trouillon, Welbl, Riedel, Gaussier, and
  Bouchard}]{trouillon_complex_2016}
Théo Trouillon, Johannes Welbl, Sebastian Riedel, Éric Gaussier, and
  Guillaume Bouchard. 2016.
\newblock \href {http://arxiv.org/abs/1606.06357} {Complex {Embeddings} for
  {Simple} {Link} {Prediction}}.
\newblock \emph{arXiv:1606.06357 [cs, stat]}.
\newblock ArXiv: 1606.06357.

\bibitem[{Xiong et~al.(2019)Xiong, Yu, Chang, Guo, and
  Wang}]{Xiong2019ImprovingQA}
Wenhan Xiong, Mo~Yu, Shiyu Chang, Xiaoxiao Guo, and William~Yang Wang. 2019.
\newblock Improving question answering over incomplete kbs with knowledge-aware
  reader.
\newblock In \emph{ACL}.

\bibitem[{Yan et~al.(2021)Yan, Li, Wang, Zhang, Daoguang, Zhang, Wu, and
  Xu}]{Yan2021LargeScaleRL}
Yuanmeng Yan, Rumei Li, Sirui Wang, Hongzhi Zhang, Zan Daoguang, Fuzheng Zhang,
  Wei Wu, and Weiran Xu. 2021.
\newblock Large-scale relation learning for question answering over knowledge
  bases with pre-trained language models.
\newblock In \emph{EMNLP}.

\bibitem[{Zhang et~al.(2018)Zhang, Dai, Kozareva, Smola, and
  Song}]{Zhang2018VariationalRF}
Yuyu Zhang, Hanjun Dai, Zornitsa Kozareva, Alex Smola, and Le~Song. 2018.
\newblock Variational reasoning for question answering with knowledge graph.
\newblock In \emph{AAAI}.

\end{thebibliography}
\bibliographystyle{acl_natbib}

\appendix

% \section{Example Appendix}
% \label{sec:appendix}

% This is an appendix.

\end{document}